%% file: main.tex
\documentclass[10pt,twocolumn,letterpaper]{article}

\usepackage[pagenumbers]{iccv} % To force page numbers, e.g. for an arXiv version


\definecolor{iccvblue}{rgb}{0.21,0.49,0.74}
\usepackage[pagebackref,breaklinks,colorlinks,allcolors=iccvblue]{hyperref}

\usepackage{multirow}
\usepackage{multicol}
\usepackage{pifont}
\usepackage{amsmath} % 加入数学宏包
\usepackage{algorithm,algorithmic}
\usepackage{xcolor}

\def\paperID{5338} % *** Enter the Paper ID here
\def\confName{ICCV}
\def\confYear{2025}

\title{On Data Synthesis and Post-training for Visual Abstract Reasoning
}

\author{
	Ke Zhu$^{1,2*}$ \quad
	Yu Wang$^2$\thanks{Equal Contributions} \quad
    Jiangjiang Liu$^2$ \quad
	Qunyi Xie$^2$ \quad
	Shanshan Liu$^2$ \quad\\
    Gang Zhang$^2$ \quad \\
$^1$Nanjing University \quad
$^2$Baidu VIS \\
{\tt\small zhuk@lamda.nju.edu.cn, \{wangyu106,liujiangjiang\}@baidu.com}
}

\begin{document}
\maketitle

\begin{abstract}
This paper is a pioneering work attempting to address abstract visual reasoning (AVR) problems for large vision-language models (VLMs). We make a common LLaVA-NeXT 7B model capable of perceiving and reasoning about specific AVR problems, surpassing both open-sourced (e.g., Qwen-2-VL-72B) and closed-sourced powerful VLMs (e.g., GPT-4o) with significant margin. This is a great breakthrough since almost all previous VLMs fail or show nearly random performance on representative AVR benchmarks. Our key success is our innovative data synthesis and post-training process, aiming to fully relieve the task difficulty and elicit the model to learn, step by step. Our 7B model is also shown to be behave well on AVR without sacrificing common multimodal comprehension abilities. We hope our paper could serve as an early effort in this area and would inspire further research in abstract visual reasoning.
\end{abstract}

\section{Introduction}
\label{sec:intro}
Large Vision-Language Models (VLMs) are now equipped with advanced multimodal reasoning ability due to great efforts in large-scale image-text joint pretraining~\cite{LLM_BLIP2,LLM_MiniGPT4} and task-specific supervised finetuning~\cite{LLaVa1.5,Qwen2-VL}. Such VLMs are capable of perceiving~\cite{LLaVa1.5} and reasoning~\cite{Qwen2.5-VL} about image content, as well as making decisions~\cite{reasoning-agent}.

\begin{figure}
	\centering

    \begin{subfigure}{0.9\linewidth}
		\includegraphics[width=0.95\linewidth]{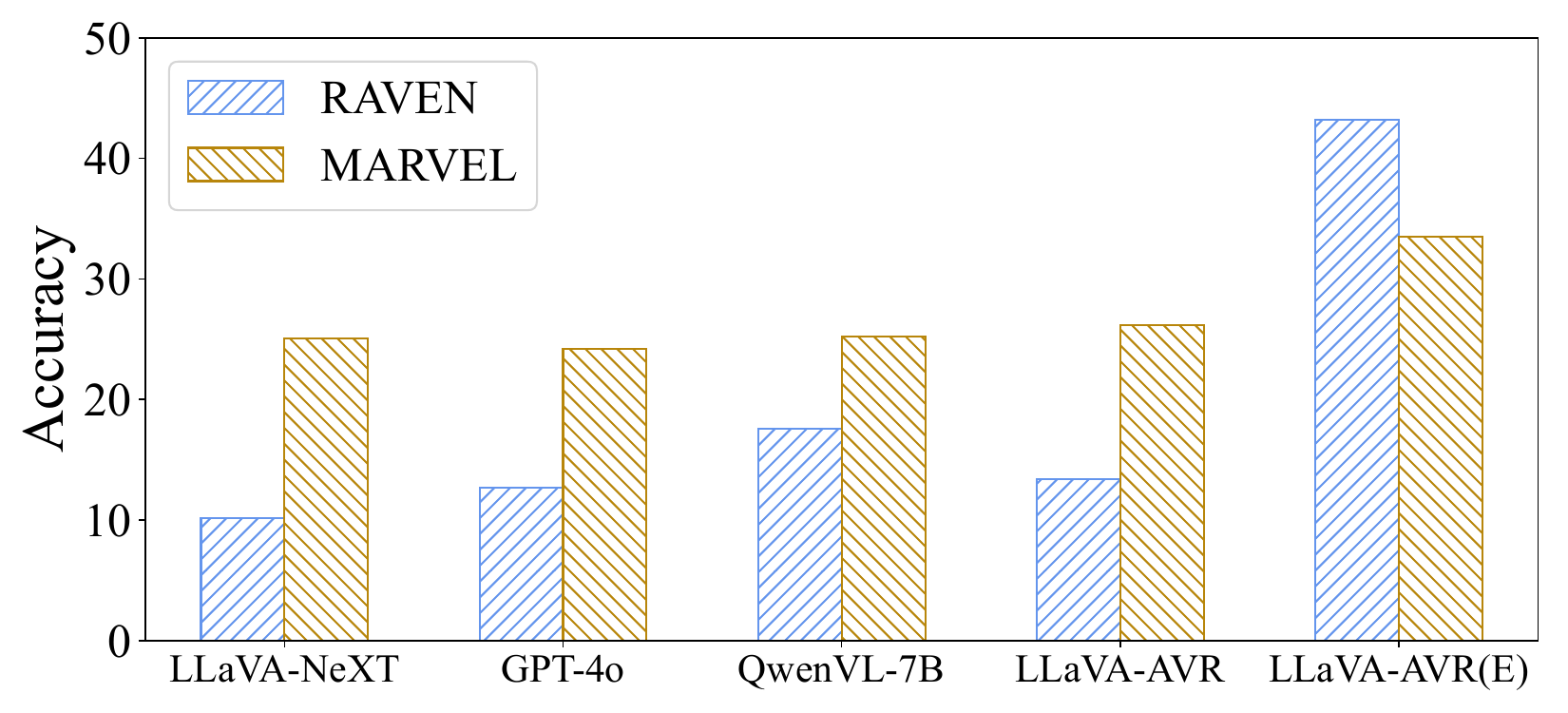}
		\caption{Results on AVR with different datasets training.}
		\label{fig:figre1-dataset-trial-1}
	\end{subfigure}
	\begin{subfigure}{0.9\linewidth}
		\includegraphics[width=0.95\linewidth]{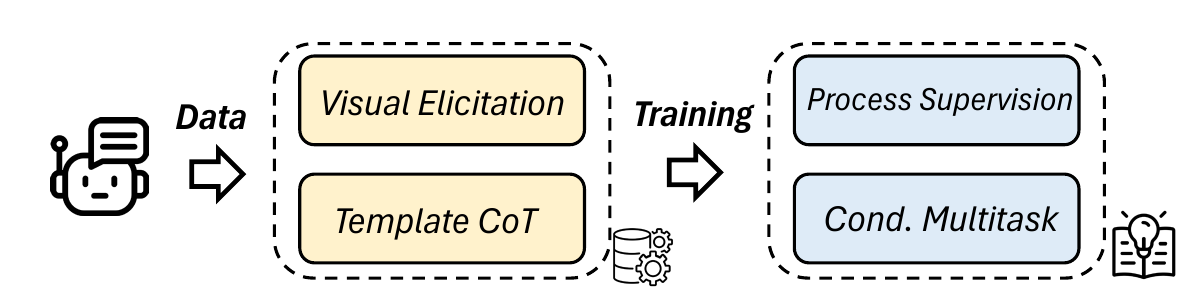}
    	\caption{Our overall elicitation process during data and training.}
		\label{fig:figure-1-pipeline}
    \end{subfigure} 

    	\caption{Fig.~\ref{fig:figre1-dataset-trial-1}: evaluation results on AVR benchmarks RAVEN~\cite{raven} and MARVEL~\cite{marvel}. LLaVA-AVR is trained with our naively collected data with original label. LLaVA-AVR(E) means we \emph{E}liciate the model to learn using our strategy shown in Fig.~\ref{fig:figure-1-pipeline}.}
	\label{fig:figre1-dataset-trial}
\end{figure}

\begin{figure*}
	\centering
	\includegraphics[width=0.92\linewidth]{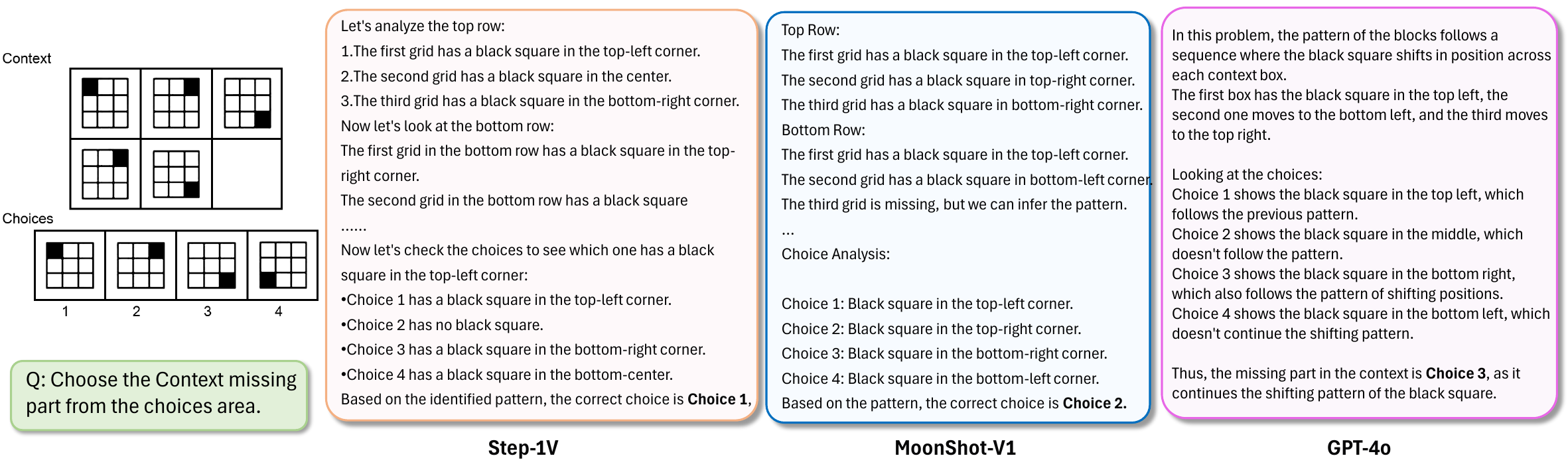}
    	\caption{The produced Chain-of-thought (CoT) by three different advanced model Step-1V~\cite{step}, MoonShot-V1~\cite{kimi} and GPT-4o. The left shown image quiz is randomly sampled from MARVEL test dataset~\cite{marvel}. The correct choice for this puzzle is 4.}
	\label{fig:compare-model-cases}
\end{figure*}

Abstract visual reasoning (AVR) recently attracts much attention in both academic and industry. On one hand, previous studies all found current VLMs' insufficiency in such scenarios (\cf Fig.~\ref{fig:figre1-dataset-trial}-\ref{fig:compare-model-cases}), pointing out the key obstacles lies in the lack of \emph{perception} and \emph{reasoning} ability. On the other hand, properly solving such tasks is highly practical, as AVR is very much relavant to education~\cite{ai4edu,the-curious}. So far as we know, very few works have truly started in this field.

This paper makes the first attempt trying to \emph{solve} the AVR tasks. Our main strategy is to \emph{elicit} the model's learning to reduce task difficulty, achieved through both data synthesis and training strategy aspects.

We first conduct an empirical study in Fig.~\ref{fig:figre1-dataset-trial}. Here we collect AVR reasoning related corpus and its tagged labels (ususally with short and direct answers), covering both RAVEN~\cite{raven} and MARVEL~\cite{marvel} domain data. Then we directly feed them into LLaVA-NeXT in a single-stage training. The model after trained is called `LLaVA-AVR'. As shown in Fig.~\ref{fig:figre1-dataset-trial-1}, \emph{naively} data sythesis and training lead to only minor improvement to the baseline, still lagging far behind more powerful models. Then, a natural question arise: \emph{when the data are available, how can we better elicit the model to learn, step by step?} More specifically, \emph{how can we better optimize data synthesis and post-training to overcome obstacles in AVR perception and reasoning?} 
% \begin{enumerate}[label=\textbullet]
%     \item \textbf{Data scarcity.} How to collect suitable visual reasoning data and more importantly, how to properly construct conversation \emph{suitable for AVR problems}?
%     \item \textbf{Training Strategy.} How to (perhaps) progressively train VLMs to help them better perceive and reason?
% \end{enumerate}

To address this, we adopt a structured strategy (\cf Fig.~\ref{fig:figure-1-pipeline}) to progressively guide the model: 1) Data: we automatically collect 32k AVR related images, and construct visual perception and reasoning chain-of-thought (CoT) data. Then, visual elicitation and templated-based CoT are adopted to facilitate faster learning without hacking (\cf Fig.~\ref{fig:ablation-elicitation-qa}). 2) Training: process-level supervision and conditional multi-task learning are utilized during training procedure to stimulate model's potential (\cf Fig.~\ref{fig:ablate-process-control-vis} and Table~\ref{tab:ablate-multi-task}).

With these weapons at hand, our post-trained LLaVA-NeXT-7B model start to \emph{perceive} and \emph{reason} in AVR problems, achieving a pioneer score on most representative AVR benchmarks that requires complex visual reasoning abilities (\cf LLaVA-AVR(E) in Fig.~\ref{fig:figre1-dataset-trial-1}). This overcomes the long-standing barrier where most advanced VLMs (e.g., GPT-4o-mini) previously exhibited nearly random performance. 

Finally, we provide solid experiments and quantitative visualizations to verify the effectiveness of the proposed innovations in data and training pipeline (\cf Table~\ref{tab:ablate-overall}). Each component proves to be indispensable and collectively ensure optimal model performance. Ablations further demonstrate that incorporating this AVR ability \emph{does not} compromise the model's original comprehension skills. We hope that our early exploration in the AVR domain could shed light on later advancements in multi-modal reasoning.

Overall, our contributions are:
\begin{enumerate}[label=\textbullet]
    \item We made an initial attempt in AVR domain, trying \emph{overcome} the key obstacles inherent in the task.
    \item We introduce innovations in the data and training pipeline, aiming to alleviate task difficulty while simultaneously eliciting the model's learning process.
    \item Our LLaVA-AVR-7B, is able to perceive and reason AVR related problems, surpassing current advanced large VLMs (e.g., GPT-4o) with non-trivial margins. 
\end{enumerate}

\section{Related Work}
\subsection{Large Vision-Language Models}
Large vision-language models (VLMs)~\cite{seva,nsft} are capable of handling multiple vision tasks like visual question answering~\cite{TextVQA}, visual grounding~\cite{cogvlm} and reasoning~\cite{MMVet}. Among them, two core abilities are essential: visual perception and the reasoning skills resided in the large language models (LLMs). Recent advanced VLMs, like Qwen series~\cite{Qwen2-VL,Qwen,Qwen2.5-VL}, GPT-4o~\cite{ShareGPT4V}, Step-1V~\cite{step}, also manifest chat ability with superior user experience. These important achievements rely on diverse image-text data source during pretraining and supervised finetuning (SFT) stage, and current focus in multmodal LLMs has gradually changed from model architectures design~\cite{LLM_Flamingo,LLM_BLIP2} to higher data~\cite{ShareGPT4V} and more efficient algorithms~\cite{dpo-or-ppo}. 

\subsection{Reasoning in LLMs and VLMs}
Reasoning techniques in LLMs has become mature in publicity~\cite{cot,pot,ChatGPT}. Representative methods to elicit LLM reasoning are chain-of-thought (CoT), program-of-thought (PoT), helping model to generate intermediate steps before drawing a conclusion. These techniques have greatly benefited LLMs, especially those with great intelligence~\cite{ChatGPT} (e.g, $>$100B). The concept of multimodal reasoning, is perhaps more general, including both entity-based reasoning (e.g, common visual question answering~\cite{gqa,sqa}) and symbolic reasoning like math or geometry reasoning~\cite{math-llava,mathvision,geoqa}. An undeniable fact is that LLM reasoning has greatly facilicated multmodal domain~\cite{mc-cot,multimodal-cot}. Besides, there is also a trend in multimodal reasoning to involve more advanced techniques like AI agent~\cite{reasoning-agent} and RAG~\cite{image-of-thought}.

\subsection{Abstract Visual Reasoning}
Abstract visual reasoning (short for AVR) has recently attracted much attention. The layout of such problems usually follows the Raven Progressive Matrix (RPM)~\cite{rpm}, and the ultimate goal in AVR is to deduce the missing pattern based on observed pattern and rule across rows or columns. Previous researches~\cite{the-curious,marvel,puzzle-vqa} focusing on AVR mostly try to \emph{analyze} and evaluate the difficulty lies within this settings, pointing out the core obstacles is the lack of perception and reasoning ability in current large vision-language models. To the best of our knowledge, this paper is the first work that attempts to \emph{solve} this AVR problem. The core component is to fully relieve the task difficulty, trying to help model to perceive and to reason, step by step.

\begin{figure*}
	\centering
	\includegraphics[width=0.92\linewidth]{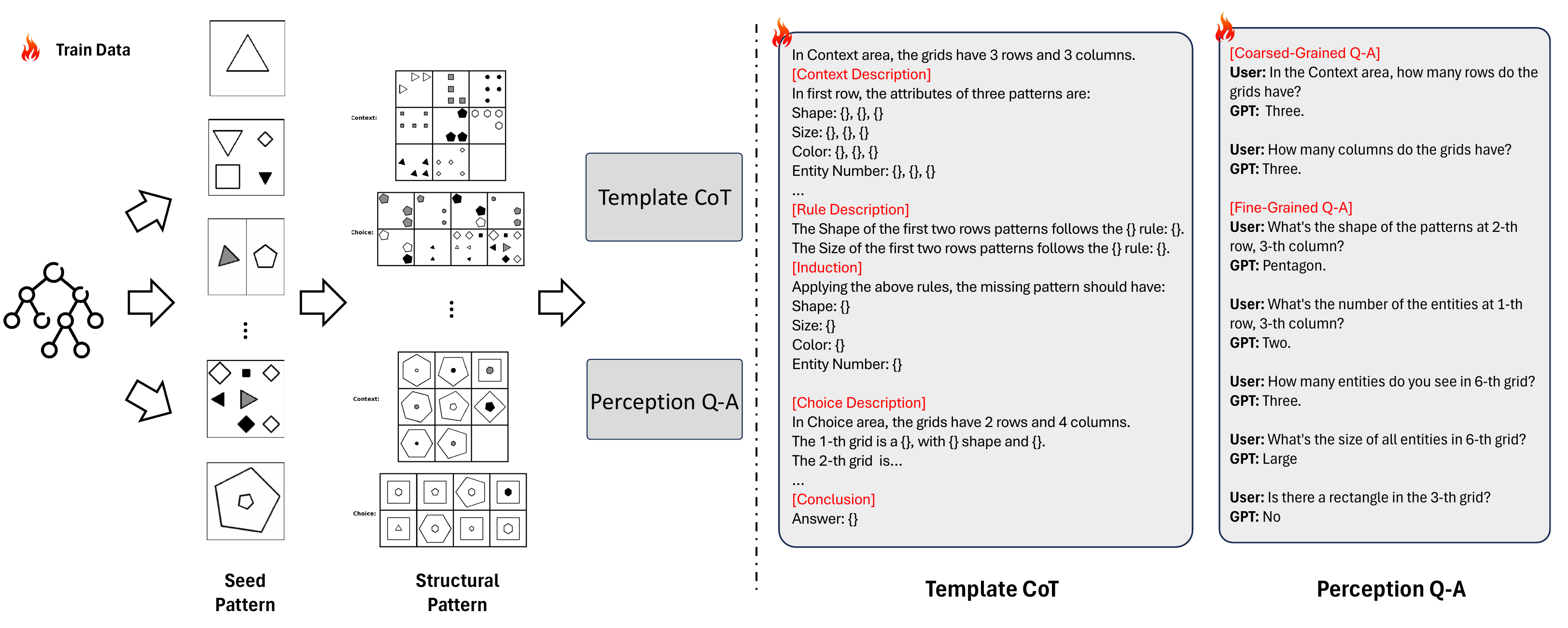}
    	\caption{Our data generation pipeline for the regular puzzle. We first choose seven different seed pattern from the initial tree, then apply the sampled rule to generate the whole mage (structural pattern). We then generate the template-based chain-of-thought and perception question-answer based on the information stored in previous process. The whole process do not involve any LLM or human effort.}
	\label{fig:method-main}
\end{figure*}

\begin{figure*}
	\centering
	\includegraphics[width=0.92\linewidth]{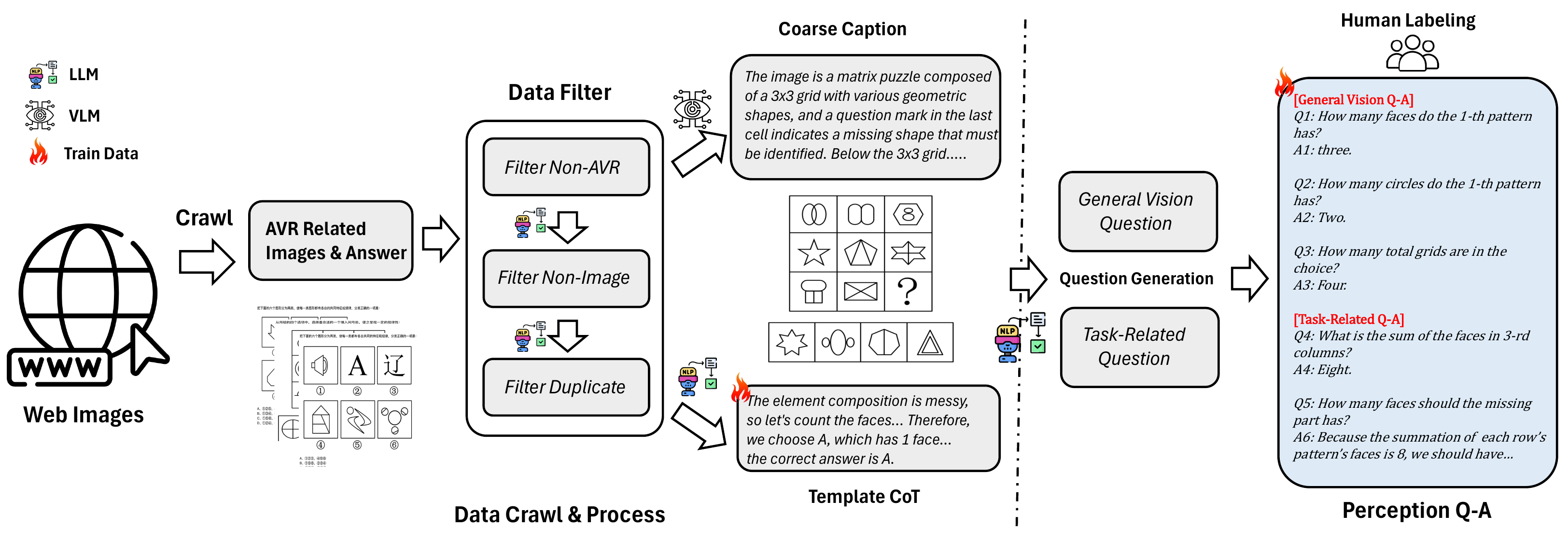}
    	\caption{Our data generation pipeline for the non-regular puzzle crawled from the CCSE website. We totally crawled about 8k data, with 4k remaining after data filtering process. We then generate coarse caption and reformat the original answer into template CoT, both of which go through an LLM to obtain specific questions for each images. Finally, we use human labor to manually annotate these questions.}
	\label{fig:method-marvel}
\end{figure*}

\section{Method}
We will first introduce basics of VLMs. Then move onto our innovative pipeline in aspects of data and training strategy. 

\subsection{Architecture}
We mainly adopt LLaVA-NeXT as our vision language models. Specifically, an image $I$ first goes through an image processor $T$ (including both the ViT and MLP layers~\cite{LLaVa1.5}) to obtain the image embeddings $v: v = T(I)$, which are combined with the question prompt $q$ ($\boldsymbol{x} = (q;I)$), and are sent into an LLM that generates the next token in order:
\begin{equation}
    \pi_{\theta}(\boldsymbol{y}|\boldsymbol{x}) = \prod_{i=1}^{L} \pi_{\theta}(y_i|y_{<i}, \boldsymbol{x})\,.
    \label{eq:sft-sequential}
\end{equation}
This generation process are optimized with a cross entropy loss (SFT loss) per token, demonstrated as follows:
\begin{align}
    \mathcal{L_\text{sft}} (\boldsymbol{y}) = -\sum_{i=1}^{L} \log \pi_{\theta}(y_i | y_{<i}, \boldsymbol{x}) \,. \label{eq:sft-loss-2} 
\end{align}

\begin{table*}
	% \footnotesize
    \small

	\centering
	\begin{tabular}{cccccccc}
		\toprule[1pt]
   \multirow{2}{*}{id} & \multicolumn{2}{c}{Data Strategy} & \multicolumn{2}{c}{Training Strategy} &  \multirow{2}{*}{Stages}  & \multirow{2}{*}{Reasoning acc} & \multirow{2}{*}{Percept. acc}\\
        
	 & Visual Elicitation & Template CoT & Local Sup. & Cond. Multi-Task & \\
    
     \midrule
   0 & \textcolor{lightgray}{--} & \textcolor{lightgray}{--} & \textcolor{lightgray}{--} & \textcolor{lightgray}{--} & N/A & 11.2 & N/A  \\
   1 & \ding{51} & & & & stage-1 & N/A & 95.2\\
   2 & \ding{51} & & & & stage-1 \& 2 & 60.2 & 79.6 \\
   3 & \ding{51} & \ding{51} & & & stage-1 \& 2 & 62.8 & 86.2\\
   4 & \ding{51} & \ding{51} & \ding{51} & & stage-1 \& 2 & 72.1 & 95.1\\
   5 & \ding{51} & \ding{51} & \ding{51} & \ding{51} & stage-1 \& 2 & \textbf{82.7} & \textbf{96.2} \\
   % 3 & & &  &  & stage-2 & 76.5 & 75.5 \\
   % 2 & & & \ding{51} &  & stage-2 & 81.8 & 86.5\\
   % 3 & & & \ding{51} & \ding{51} & stage-1,2 & \textbf{83.0} & \textbf{90.2}\\
    \bottomrule
	\end{tabular}
\caption{A full illustration of the proposed innovative Data synthesis approach and Training strategy. The evaluation datasets are chosen as RAVEN~\cite{raven} since its metrics are easier to quantify (\cf appendix). The first line refers to the LLaVA-NeXT-7B models.}
\label{tab:ablate-overall}
\end{table*}

\subsection{Data Synthesis}
\label{sec:data-synthesis}
We collected two source of data, covering both regular pattern puzzle and non-regular puzzle. For each data source, we manually filter the test related images existed in RAVEN and MARVEL to prevent hacking. Generally, we synthesized perception question-answering and template-based CoT for each type of puzzle, which are utilized for model training. Please refer to Fig.~\ref{tab:main-raven}-\ref{tab:main-marvel} for the process illustration, and confer `Dataset' in Table~\ref{tab:dataset-stage-train} for an overall look.

\begin{figure}
	\centering
	\includegraphics[width=0.92\linewidth]{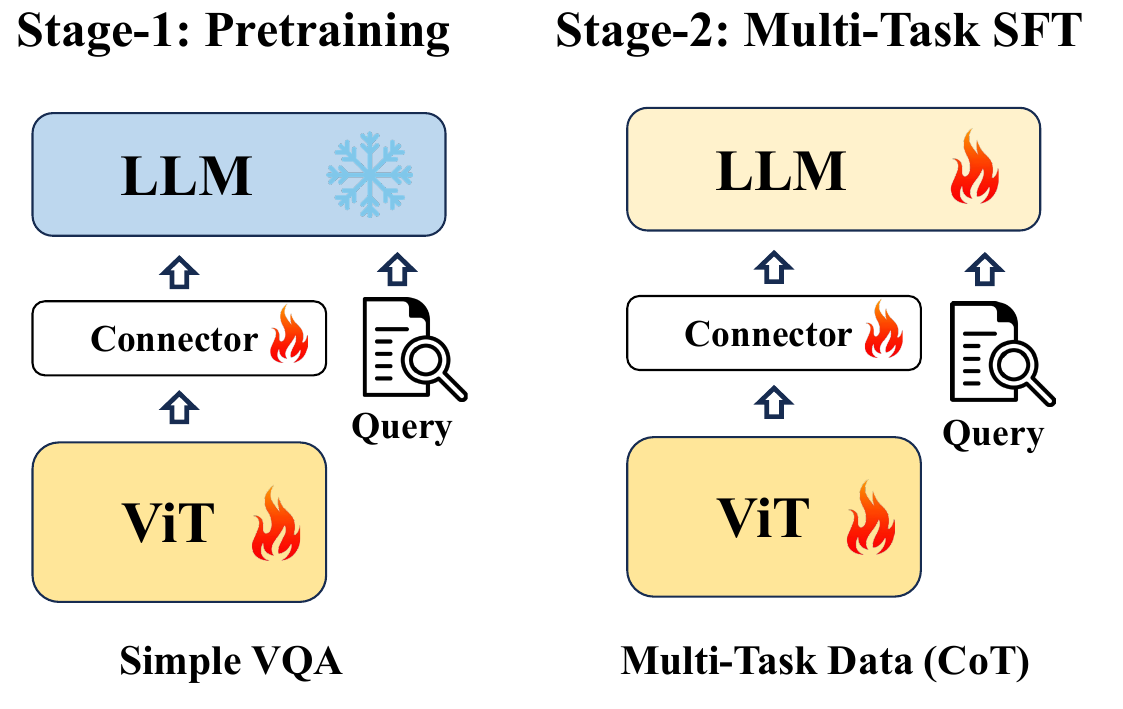}
    	\caption{The training pipeline of our model LLaVA-AVR-7B, including Pretraining stage with short perception VQA, and Multi-task Supervised finetuning with both perception VQA and long CoT. The stage-1 model are all initialized with LLaVA-NeXT-7B.}
	\label{fig:method-stage}
\end{figure}

\textbf{Regular puzzle.} This kind of data has limited attributes with fixed pattern variation, which quite resemble RAVEN's distribution, and its generation process is \emph{fully automatic}. We use the Attributed Stochastic Image Grammar Tool (A-SIG)\cite{a-sig} for data generation. Specifically, we first sample a predefined A-SIG sentence and a variation rule, and renders the seed pattern image. Then we apply the rule to the seed pattern that sequentially generates the whole structural puzzle. Each puzzle's information (pattern and variation rule) are pre-recorded during the generation process, which are utilized to form the chain-of-thought and automatically generate the perception question-answers. During question-answering (Q-A) process, we adopt visual elicitation prompting in perception Q-A, and use a template CoT to relieve the learning difficulty, which are named as RAVAE-VQA and RAVEN-CoT, respectively. The whole generation process is demonstrated in Fig.~\ref{tab:main-raven}.  For RAVEN evaluation, we generate a batch of test data in parallel, but a totally different seed, to guarantee there is non-overlap with our generated training data.

\textbf{Non-regular puzzle.} This kind of data generation is more complex, since the patterns in such puzzle is irregular and sometimes in a mass. Thus, there are almost no available annotations for visual perception Q-A or chain-of-thought reasoning. Inspired by previous researches~\cite{marvel,the-curious}, we obtain relavant sources from web and annotate them in a semi-supervised manner. Specifically, we first crawl images from China Civil Service Examination (CCSE) website, obtaining the initial raw images and the originally attached answers (short chain-of-thought). We then conduct automatic filtering to make sure the remaining corpus are all unique and only contain AVR images. Based on these image-answer pairs, we utilize large VLMs Qwen-2-VL-72b-AWQ to generate the coarse image caption, and use LLM to convert the original answer to a specific templated CoT format (called CCSE-CoT). Next, we generate general questions (CCSE-VQA) and Task-Related questions (called CCSE-TRVQA) for each images, based on captions and CoT answers, which are finally labeled by human labor. The overall process are clearly demonstrated in Fig.~\ref{tab:main-marvel}. We verify through ablations that \emph{human annotation} is quite essential.

\subsection{Training Strategy}
\label{sec:training}
Firstly, we warm up the vision encoder to help the model to recognize basic AVR patterns. We choose the simle perception Q-A, including RAVEN-VQA and CCSE-VQA to train the vision-encoder and the MLP adapter. We do not include task-related Q-A in CCSE since the answer's length and format do not comply with the frozen LLM's output style, which requires unfreezing LLM.

In stage-2 reasoning process, We mainly adopt two innovative training strategies, to elicit the model to perceive and to reason, in a better way. 

\textbf{Process level supervision.} This concept derives from the process reward model in reinforcement learning~\cite{deepseekmath}. Specifically, during stage-2, we involve all perception VQA adopted in stage-1 training (\cf Table~\ref{tab:dataset-stage-train}) to guarantee the local correctness for the chain-of-though reasoning process.

\textbf{Conditional multi-task learning.} This kind of strategy is much more directly and are inspired by previouse researches~\cite{sft-nlp-multitask}. The mixture of all CoT data naturally forms a multi-task format if we regard each sub seed pattern in Fig.~\ref{tab:main-raven} as a sub-task. For each task in the constructed RAVEN and CCSE data, we add a special sentence in each CoT's \emph{taget labels} to make the image content more easily distinguishable. For example, we add one sentence at the beginning of RAVEN-CoT and CCSE-CoT, respectively:

RAVEN:\texttt{ This is a regular puzzle. The grid pattern is a [xxx] style.}

CCSE:\texttt{ This is a non-regular puzzle.}

A general results can be found in Table~\ref{tab:ablate-overall}, where all data strategy (Visual Elicitation, Template CoT) and training techniques (Local Supervision and Conditional Multi-Task) are listed. As shown in the Table, pure baseline (id 0) behaves poorly on the RAVEN reasoning and perception accuracy. When we adopt visual elicitation training, the perception accuracy has seen a rapid growth. The template-based CoT also helps model to reason. In terms of training strategy, both local supervision and Conditional Multi-Task learning helps the reasoning and perception ability. Overally, we obtain a model of 82.7 reasoning ability and 96.2 perception accuracy.

\section{Experiments}
In this section, we will first provide the training settings, including our synthesized data, the evaluation dataset and the training details. Then we provide our main experimental results. Finally, fruitful of ablations are provided.

\subsection{Training Settings}

\textbf{Synthesized data.} Our synthesized data contains two part. One is the regular puzzle, resembling the RAVEN's distribution. Specifically, we construct 4k VQA and 4k CoT for each seed pattern (total 7 different seed pattern, same as RAVEN), forming a total of 28k VQA and CoT data. Note that during this data generation process, we manually prune the variation rule to make the pattern attributes more simple (\cf appendix for more details). For data crawled from CCSE, we obtain about 4k data after the filtering process, and constructed 4k VQA, 4k task-related VQA and 4k CoT, respectively. The LLM and VLM used during construction are GPT-4 and Qwen-72B-AWQ~\cite{Qwen2-VL}. Please refer to Table~\ref{tab:dataset-stage-train} for more details. We also manually exclude all data that exists in the evaluation data (RAVEN and MARVEL test set) to make the experiment fair.

\textbf{Evaluation dataset.} The evaluation dataset are mainly RAVEN~\cite{raven} and MARVEL~\cite{marvel}. Following RAVEN original settings~\cite{raven}, its evaluation dataset are generated using A-SIG~\cite{a-sig}, with the same pruned rule described above. There are total 7 seed pattern or subtasks in RAVEN, namely Center, Grid-Two (G-2), Grid-Three (G-3), Left-Right (L-R), Up-Down (U-D), Out-InCenter (O-IC) and Out-InGrid (O-IG). The MARVEL dataset contains 770 images, covering six different pattern types, namely Temporary-Movement (T-M), Spatial-Relation (S-R), Quantitle (Q-T), 2D-Geometric (2D) and 3D Geometric (3D). We use the short name to represent each sub-task. 

\textbf{Training details.} We use LLaVA-NeXT-7B as our base VLMs and continually train it using our synthetic data and the proposed training strategies. Specifically, we use DeepSpeed framework and ZeRO-3 for better optimization. The learning rate and batch size are set as 2e-6 and 4, respectively. during post-training, we first train the vision-encoder and MLP in stage-1, using RAVEN-VQA-28k and CCSE-VQA-4k. Then we totally unfreeze all the model, and train it using all the data. The model after post-training are called LLaVA-AVR-7B in the subsequent experiements.

\begin{table}
	\small
	\setlength{\tabcolsep}{3pt}
	\centering
	\begin{tabular}{ccccccc}
		\toprule[1pt]
		Config & Stage-1 & Stage-2 \\        
         \midrule
         LearningRate & 1e-5 & 1e-5 \\
         TrainingEpochs &  4 & 1\\
         BatchSize  & 2 & 4\\
         Trainable Part & vit & vit,llm\\
         Gradient Accu. & 1 & 1 \\
         Dynamic Resolution & False & False \\
         
         \midrule
        \multirow{5}{*}{Dataset} && RAVEN-VQA-28k \\
        & & RAVEN-CoT-28k \\
        & RAVEN-VQA-28k  & CCSE-VQA-4k\\
        & CCSE-VQA-4k & CCSE-TRVQA-4k\\
        & & CCSE-CoT-4k \\
        \midrule 
        Train Hours (h) & 0.5h & 1.5h\\
        \bottomrule[1pt]
	\end{tabular}
\caption{The configurations, dataset and training time cost of our Stage-1 Pretraining and Stage-2 Multi-Task SFT.}
\label{tab:dataset-stage-train}
\end{table}

\subsection{Experimental Results}
\textbf{RAVEN datasets.} We first evaluate our LLaVA-AVR-7B model on RAVEN datasets, which contains 7 sub categories. As shown in Table~\ref{tab:main-raven}, our LLaVA-AVR-7B models consistently surpass previous models in all metrics, with significant margins. In the closed-source models, GPT-4o-mini, Step-1V and Moonshot-V1 almost show random performance (around 12.5\%). Among all open-source model, Qwen-2-VL turns out to be the most powerful, showing significant advantage over others. If we inspect each tasks accuracy, we will find the most difficult ones is the `G-3' settings, where the objects size is the smallest. This indicates the lack of fine-grained ability for current VLMs in~\cite{cogvlm}.

\begin{table*}
	% \small
	\centering
	\begin{tabular}{cccccccccc}
		\toprule[1pt]
		Model  & Accuracy & \texttt{Center} & \texttt{G-2} & \texttt{G-3} & \texttt{L-R} & \texttt{U-D} & \texttt{O-IC} & \texttt{O-IG}\\
        \midrule
        \textcolor{lightgray}{\small\textit{open-source model}} \\
        InstructBLIP-7B~\cite{InstructBLIP}  & 9.7 & 14.5 & 10.2 & 2.8 & 12.3 & 15.8 & 8.2 & 3.9  \\
        LLaVA-1.5-13B~\cite{LLaVa1.5} & 10.3 & 12.8 & 13.2 & 10.2 & 9.8 & 16.4 & 6.2 & 3.8 \\
        LLaVA-NeXT-7B~\cite{llava-next} & 10.2 & 13.2 & 12.1 & 9.3 & 11.5 & 17.2 & 5.7 & 2.6\\
        Qwen-2-VL-7B~\cite{Qwen2-VL}  & 17.5 & 33.8 & 20.9 & 15.5 & 5.2 & 14.3 & 18.8 & 14.3 \\
        Qwen-2-VL-72B~\cite{Qwen2.5-VL} & 33.6 & 90.2 & 32.2 & 26.4 & 5.7 & 16.6 & 41.6 & 22.4 \\
        \midrule
        \textcolor{lightgray}{\small\textit{closed-source model}} \\
        GPT-4o-mini & 12.7 & 20.5 & 15.2 & 11.2 & 7.8 & 9.3 & 10.9 & 4.8\\
        Step-1V-8k & 11.1 & 14.3 & 10.8 & 9.5 & 14.2 & 11.9 & 11.9 & 5.8\\
        Moonshot-V1 & 14.2 & 23.8 & 14.2 & 9.5 & 14.2 & 19.1 & 4.8 & 3.2 \\
        \midrule
        LLaVA-AVR-7B  & \textbf{82.7} & \textbf{98.2} & \textbf{68.2} & \textbf{66.2} & \textbf{96.5} & \textbf{97.8} & \textbf{94.2} & \textbf{58.2}\\
        \bottomrule[1pt]
	\end{tabular}
\caption{Evaluated reasoning results on RAVEN~\cite{raven}. We evaluate five open-source models and three advanced closed source models. We also report the per sub-task's accuracy (7 in total) in the table. Our LLaVA-AVR-7B consistently surpass them in the listed metrics.}
\label{tab:main-raven}
\end{table*}

\begin{table*}
	% \small
	\centering
	\begin{tabular}{cccccccccc}
		\toprule[1pt]
		Model  & Percept. acc  & Reasoning acc & \texttt{T-M} & \texttt{S-R} & \texttt{Q-T} & \texttt{M-T} & \texttt{2D} & \texttt{3D}\\
        \midrule
        \textcolor{lightgray}{\small\textit{open-source model}} \\
        InstructBLIP-7B~\cite{InstructBLIP} & 41.5 & 25.3 & 25.7 & 21.7 & 24.6 & 29.7 & 23.6 & 25.0\\
        LLaVA-1.5-13B~\cite{LLaVa1.5} & 45.1 & 25.4  & 28.6 & 30.0& 19.6 & 26.1 &29.2 & 20.0 \\
        LLaVA-NeXT-7B~\cite{llava-next} & 46.2 & 25.4 & 21.9 & 27.5 & 25.8 & 26.1 & 25.8 & 20.0 \\
        Qwen-2-VL-7B~\cite{Qwen2-VL} & 54.2 & 25.2 & 25.7 & 21.7 & 24.6 & 29.7 & 23.6 & 25.0\\
        Qwen-2-VL-72B~\cite{Qwen2-VL} & 70.1 & 26.8 & 26.6 & 24.2 & 29.2 & 27.9 & 25.0 & 25.0 \\
        \midrule
        \textcolor{lightgray}{\small\textit{closed-source model}} \\
        GPT-4o-mini & 50.1 & 24.2 & 22.8 & 25.8 & 25.0 & 21.2 & 26.7 & 20.0\\
        Step-1V-8k  & 73.8 & 26.6 & 28.6 & 35.8 & 22.5 & 24.8 & 25.0 & 35.0\\
        Moonshot-V1 & 59.9 & 24.4 & 23.8 & 24.2 & 25.4 & 20.0 & 29.2 & 25.0\\
        \midrule
        LLaVA-AVR-7B & \textbf{75.5} & \textbf{35.7} & \textbf{37.1} & \textbf{30.0} & \textbf{35.0} & \textbf{35.7} & \textbf{42.5} & \textbf{35.0} \\ 
 
        \bottomrule[1pt]
	\end{tabular}
\caption{Results on the MARVEL~\cite{marvel} datasets. We evaluate five open-source models (e.g., Qwen-2-VL series) and three powerful closed source models (e.g., GPT-4o-mini). With our training pipeline, our LLaVA-AVR-7B surpass previous state-of-the-art, especially on the perception accuracy. We also report the accuracy of each six sub-category in this table.}
\label{tab:main-marvel}
\end{table*}

\textbf{MARVEL datasets.} We then evaluate our model on MARVEL~\cite{marvel} datasets. As seen in Table~\ref{tab:main-marvel}, our LLaVA-AVR-7B model achieves the overall best accuracy on the perception and reasoning metrics. Specifically, our model surpasses Qwen-2-VL-72B and GPT-4o-mini by 8.9 and 11.5 point in reasoning, respectively. However, this dataset is more challenging, since even with our carefully designed human labeling, the reasoning accuracy do not increase as fast as that in RAVEN dataset (Note that the human level is only about 68\% reasoning accuracy shown in~\cite{marvel}). One possible reason is that our annotation do not contain \emph{all possible} attribute and pattern as that in RAVEN. Since the different pattern in CCSE is much more diverse and difficult to annotate all of them (\cf appendix), we thus \emph{sincerely call on researchers to include more quality annotations that could fully solve this tasks}.

\begin{figure*}
	\centering
    \begin{subfigure}{0.67\linewidth}
		\includegraphics[width=0.95\linewidth]{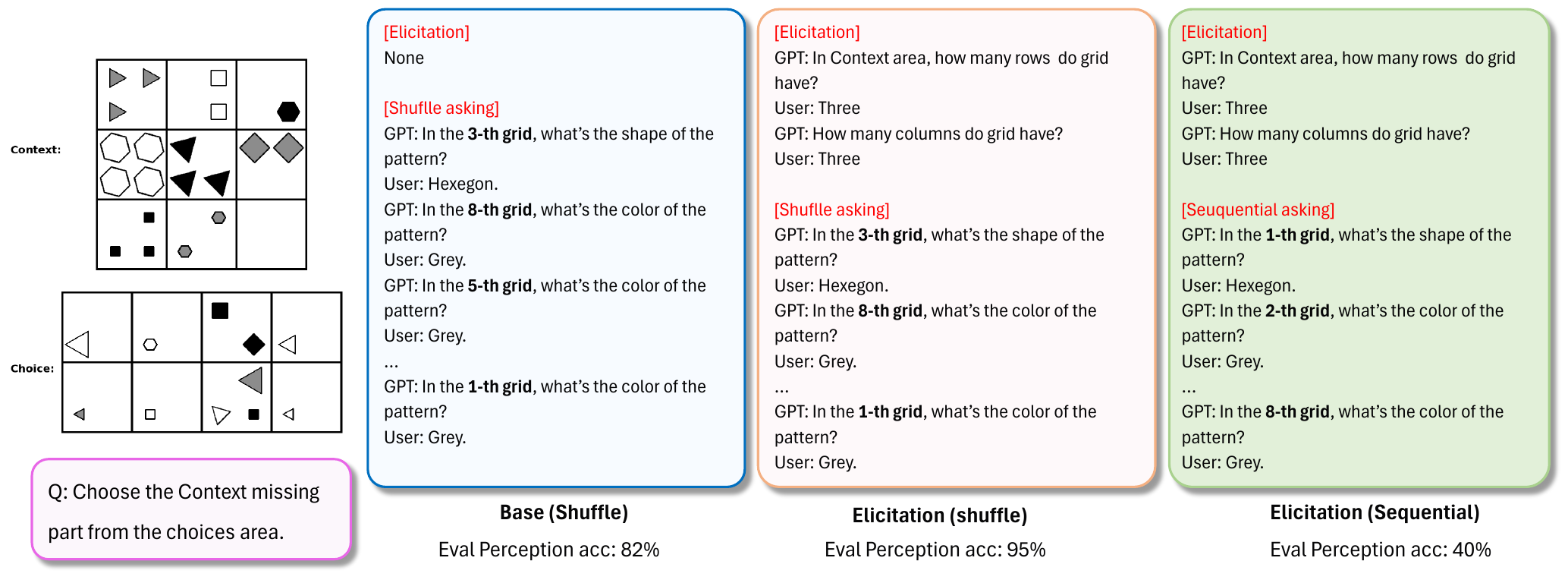}
		\caption{Three ways to construct Q-A and its perception accuracy}
		\label{fig:ablation-elicitation-qa-answer}
	\end{subfigure}
	\begin{subfigure}{0.3\linewidth}
		\includegraphics[width=0.95\linewidth]{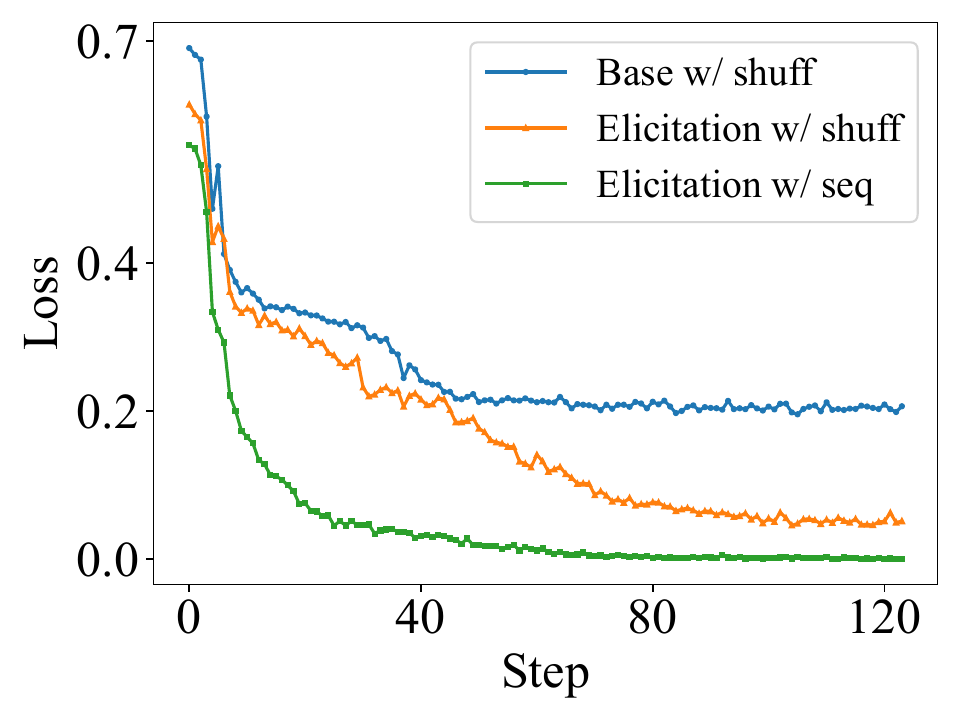}
    	\caption{Corresponding loss curve.}
		\label{fig:ablation-elicitation-qa-curve}
    \end{subfigure} 

	\caption{Ablation on the Visual Elicitation. Elicitation means we first force model to answer the puzzle structure before moving to fine-grained details. Shuffle means the fine-grained question are asked in a shuffled grid order. The evaluation perception accuracy are attached in~\ref{fig:ablation-elicitation-qa-answer}, and the corresponding training loss curve are shown in~\ref{fig:ablation-elicitation-qa-curve}. This figure is best to be viewed in color.}
	\label{fig:ablation-elicitation-qa}
\end{figure*}

\begin{table*}
	\small

	\centering
	\begin{tabular}{cccccccccc}
		\toprule[1pt]
		   Train Strategy & Epoch & Accuracy  & \texttt{Center} & \texttt{G-2} & \texttt{G-3} & \texttt{L-R} & \texttt{U-D} & \texttt{O-IC} & \texttt{O-IG}\\
        \midrule
        Single-Task & 1 & 70.6 & 92.2 & 59.0 & 42.3 & 92.2 & 92.8 & 82.2 & 33.5\\
        Single-Task & 2 & 76.4 & 96.8 & 66.3 & 52.2 & 93.3 & 94.2 & 88.6 & 43.8\\ 
        Multi-Task & 1 & 72.1 & 97.8 & 56.2 & 60.0 & 92.8 & 96.5 & 73.3 & 28.1 \\
        Cond. Multi-Task  & 1 & \textbf{82.7} & \textbf{98.2} & \textbf{68.2} & \textbf{66.2} & \textbf{96.5} & \textbf{97.8} & \textbf{94.2} & \textbf{58.2} \\
        
        \bottomrule[1pt]
	\end{tabular}
\caption{Comparison between single/multi task. Single-Task refers to the base model (LLaVA-NeXT) respectively learns a single task (e.g., `G-2') at a time, and report its corresponding sub-task accuracy. Multi-task is the default settings where all data are jointly trained. Conditional Multi-Task means a specific classification prompt are appended to the answer (\cf Sec.~\ref{sec:training}), which is utilized in our pipeline.}
\label{tab:ablate-multi-task}
\end{table*}

\begin{table}
	% \footnotesize
    \small

	\centering
	\begin{tabular}{ccccc}
		\toprule[1pt]
	Caption Model & Reasoning Acc & Perception Acc \\
     \midrule
        GPT-4V & 28.6 & 52.1 \\
        GPT-4o-mini & 26.5 & 53.2 \\
        Qwen-2-VL-72B & 29.2 & 60.2 \\
        Human Label & \textbf{35.7} & \textbf{75.5} \\
        Human Label w/ TRQ & 32.9 & 60.2 \\
    \bottomrule
	\end{tabular}
\caption{Ablations on human labor during CCSE data construction. We compare GPT-4V, GPT-4o-mini and Qwen-2-VL-72B as alternatives for human labeling during perception data construction (\cf Table~\ref{tab:main-marvel}), and evaluate the on the MARVEL~\cite{marvel} dataset. `TRQ' means the task-related questions (\cf. Table~\ref{tab:dataset-stage-train}).}
\label{tab:ablate-human-label}
\end{table}

\begin{table*}
	% \footnotesize
    \small

	\centering
	\begin{tabular}{cccccccccc}
		\toprule[1pt]
	\multirow{2}{*}{id} &\multirow{2}{*}{Strategy} & \multirow{2}{*}{Trainable part} &   \multirow{2}{*}{Data Mixture} & \multirow{2}{*}{RAVEN} & \multicolumn{5}{c}{Multimodal Comprehension} \\
     & & & & &  MMB & SQA & GQA & MME & MMMU \\
     \midrule
    0 &\textcolor{lightgray}{baseline} & \textcolor{lightgray}{---} &\textcolor{lightgray}{---} & 11.2 & \textbf{67.2} & 71.2 & 62.2 & 1503 & 33.9 \\
    1 & Post-train & vit,llm & RAVEN-28k & \textbf{82.1} & 65.1 & 68.1 & 61.8 & 1453 & 32.2\\
    2 & Post-train & llm (LoRA) & RAVEN-28k & 78.2 & 66.9 & 70.8 &  61.6 & 1501 & 33.8\\
    3 & Post-train & vit,llm & RAVEN-28k + LN 10\% & 82.2 &  66.8 & 70.4 & \textbf{63.0} & 1505 & 33.8 \\
    4 & SFT-Stage & vit,llm & RAVEN-28k + LN full & 81.2 & 67.0 & \textbf{71.3} & 62.3 & \textbf{1511} & \textbf{34.2} \\
        \bottomrule[1pt]
	\end{tabular}
\caption{Investigation on learning new capabilities without sacrificing common multimodal comprehension abilities. RAVEN-28k means the combination of RAVEN-VQA-28k and RAVEN-CoT-28k. The baseline results is the LLaVA-NeXT-7B's results. During Post-training, we apply LoRA to prevent distribution shift (\cf id 2), the mixture of RAVEN and LLaVA-NeXT SFT 10\% data (\cf id 3). We also merge RAVEN into LLaVA-NeXT SFT stage for joint training (\cf id 4).}
\label{tab:ablate-continual-learning}
\end{table*}

\begin{figure*}
	\centering
	\includegraphics[width=0.87\linewidth]{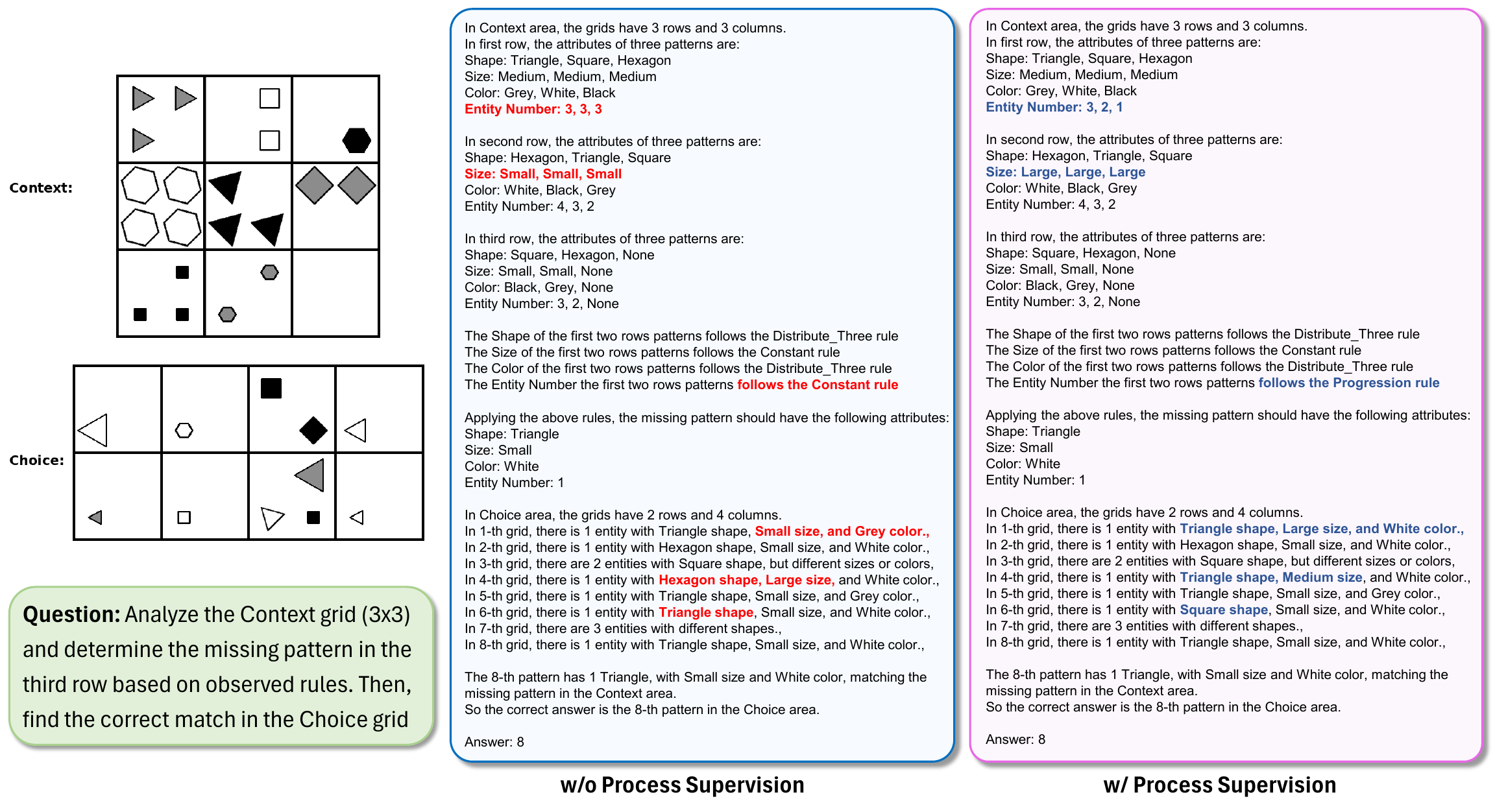}
    	\caption{The effect of applying process-level supervision (adding perception Q-A during stage-2 multi-task CoT training, \cf Table~\ref{tab:dataset-stage-train}). With proper process supervision, the local details chain-of-thought will be more correct in comparison. }
	\label{fig:ablate-process-control-vis}
\end{figure*}

\subsection{Ablations}
In this subsection, we will fully explore the effect of our component in both data and training aspects.

\textbf{Visual Elicitation.} Now we give a deeper analysis of the elicitation process in Fig.~\ref{fig:ablation-elicitation-qa}. Here we illustrate three ways to construct the visual perception questions. The `Base (Shuffle)' method means we did not involve context questions at the beggining, and directly forces the model to learn later fine-grained questions. The `Elicitation (shuffle)' is our default adopted approach, where we first force the model to answer global context question before moving to more details. `Elicitation (Sequential)' uses Elicitation at the start, but sequentially ask model fine-grained question following the grid order. As observed in Fig~\ref{fig:ablation-elicitation-qa}, after proper elicitation (compare `base' and elicitation (shuffle)), the model converges faster and achieves better perception accuracy, showing that elicitation is valid. Interestingly, the `Elicitation (sequential)' obtains the quickest convergent speed, but achieves the worst accuracy. We guess the model fail to looking at the image content during learning, but it utilizes the pattern variation rule hidden in the asking order, to answer the latter fine-grained questions. We will leave this interesting observation as future work and may visualize the vision attention score for different elicitation method. 

\textbf{Condition Multi-Task.} Now we provide a comprehensive results to show the superiority of involving the conditional signals in multi-task learning. The results can be found in Table~\ref{tab:ablate-multi-task}. Here `Single-Task' means we train different models from LLaVA-NeXT-7B for each specific sub-tasks. The Multi-task is the popular training settings in current VLMs where multiple data are directly merged. The Conditional Multi-task is our default settings, where each identifier is appended at the target label sentence (\cf Sec.~\ref{sec:training}). As shown in the table, Multi-task shows minor improvement to single task, but with the conditional signals, the model's performance significantly increase, showing the effectiveness of our strategy.

\textbf{Process level supervision.} We then visualize the effect of involving process level supervision (involving perception VQA during stage-2 training, \cf Table~\ref{tab:dataset-stage-train}), in Fig.~\ref{fig:ablate-process-control-vis}, without local control, the output chain-of-thought will sometimes incur perception error. This effect is prevented when process level supervision is involved. Although in this sampled case, the model still made the correct conclusion, we empirically verify that process level supervision will generally lead to better reasoning accuracy, as clear deomonstrated in Table~\ref{tab:ablate-overall} (compare id 3 and id 4).

\textbf{Neccessity of human labeling.} Since we involve human labor to annotate the generate question, we now provide facts to show that this procedure is indeed necessary. As can be seen in Table~\ref{tab:ablate-human-label}, utilizing GPT-4V/4o-mini or open-source Qwen-2-VL-72B will all lead to suboptimal results, mostly because that these models themselves are prone to perception mistakes in CCSE dataset (\cf Table~\ref{tab:main-marvel}). We also found in the table that involve task-related question (TRQ, \cf Table~\ref{tab:dataset-stage-train}) is necessary to achieve a decent performance, indicating that task-related annotation might be the most helpful besides simple visual perception questions. 

\textbf{Multimodal Comprehension tax.} Last but not least, we verify whether this newly involved ability in abstract visual reasoning will impair the original multimodal comprehension ability. We try four different settings (as shown in id 1-4 in Table~\ref{tab:ablate-continual-learning}). We take the generated RAVEN synthesized data for analysis for a more pure conclusion. The id 1 is our default settings where only RAVEN dataset are involved, which demonstrated decent performance on the RAVEN evaluation dataset. However, its multimodal comprehension ability are somehow lost, as shown in the MMB~\cite{MMBench}, SQA~\cite{sqa} benchmarks. Using Adapter (id 2) is more effective, but the improvement on specific domain results (on RAVEN) is limited. In comparison, using a 10\% portion of LLaVA-NeXT-738k or merge the RAVEN-28k into SFT stage will both boost the results on RAVEN without sacrificing the model's original multimodal comprehension ability. 

\section{Conclusion and Limitations}
In this paper, we advocate that the core obstacles in abstract visual reasoning lies in the data scarcity and the sub-optimal training strategy. We thus innovatively design proper data synthesis and training pipeline that fully relieves the task difficulty. We synthesized about 28k and 4k for the regular and irregular puzzle, respectively, both of which went automatic or semi-automatic labeling. With this, we successfully achieve the state-of-the-art performance in representative AVR benchmarks. We also conduct sufficient ablation to further illustrate the validity of the proposed method.

As for the limitations, we found that the reasoning performance on MARVEL is still limited (about 35.7\%). Given that human level results is only about 68\%~\cite{marvel}, we conjecture that irregular puzzle quiz is still an open problems with big challenges. We guess that more human labeling will be beneficial, but it means introducing more annotations cost. Using open-source VLM to label will be easier to scale up, but the accuracy will not be guaranteed. One possible way to totally solve complex AVR like MARVEL is to create a \emph{huge} attribute set that covers all of its varied attributes, and enlarge the training data scale. We thus call on researchers to jointly engage in AVR area (and perhaps the education domain) to explore how to better label those complicated problems with a best economical trade-off.

{
    \small
    \bibliographystyle{ieeenat_fullname}
    \bibliography{main}
}

\end{document}

%% file: main.bbl
\begin{thebibliography}{40}
\providecommand{\natexlab}[1]{#1}
\providecommand{\url}[1]{\texttt{#1}}
\expandafter\ifx\csname urlstyle\endcsname\relax
  \providecommand{\doi}[1]{doi: #1}\else
  \providecommand{\doi}{doi: \begingroup \urlstyle{rm}\Url}\fi

\bibitem[Ahrabian et~al.(2024)Ahrabian, Sourati, Sun, Zhang, Jiang, Morstatter, and Pujara]{the-curious}
Kian Ahrabian, Zhivar Sourati, Kexuan Sun, Jiarui Zhang, Yifan Jiang, Fred Morstatter, and Jay Pujara.
\newblock The curious case of nonverbal abstract reasoning with multi-modal large language models.
\newblock In \emph{First Conference on Language Modeling}, 2024.

\bibitem[Alayrac et~al.(2022)Alayrac, Donahue, Luc, Miech, Barr, Hasson, Lenc, Mensch, Millican, Reynolds, Ring, Rutherford, Cabi, Han, Gong, Samangooei, Monteiro, Menick, Borgeaud, Brock, Nematzadeh, Sharifzadeh, Binkowski, Barreira, Vinyals, Zisserman, and Simonyan]{LLM_Flamingo}
Jean{-}Baptiste Alayrac, Jeff Donahue, Pauline Luc, Antoine Miech, Iain Barr, Yana Hasson, Karel Lenc, Arthur Mensch, Katherine Millican, Malcolm Reynolds, Roman Ring, Eliza Rutherford, Serkan Cabi, Tengda Han, Zhitao Gong, Sina Samangooei, Marianne Monteiro, Jacob~L. Menick, Sebastian Borgeaud, Andy Brock, Aida Nematzadeh, Sahand Sharifzadeh, Mikolaj Binkowski, Ricardo Barreira, Oriol Vinyals, Andrew Zisserman, and Kar{\'{e}}n Simonyan.
\newblock Flamingo: a visual language model for few-shot learning.
\newblock In \emph{Advances in Neural Information Processing Systems}, 2022.

\bibitem[Bai et~al.(2023)Bai, Bai, Yang, Wang, Tan, Wang, Lin, Zhou, and Zhou]{Qwen}
Jinze Bai, Shuai Bai, Shusheng Yang, Shijie Wang, Sinan Tan, Peng Wang, Junyang Lin, Chang Zhou, and Jingren Zhou.
\newblock Qwen-vl: A versatile vision-language model for understanding, localization, text reading, and beyond.
\newblock 2023.

\bibitem[Bai et~al.(2025)Bai, Chen, Liu, Wang, Ge, Song, Dang, Wang, Wang, Tang, Zhong, Zhu, Yang, Li, Wan, Wang, Ding, Fu, Xu, Ye, Zhang, Xie, Cheng, Zhang, Yang, Xu, and Lin]{Qwen2.5-VL}
Shuai Bai, Keqin Chen, Xuejing Liu, Jialin Wang, Wenbin Ge, Sibo Song, Kai Dang, Peng Wang, Shijie Wang, Jun Tang, Humen Zhong, Yuanzhi Zhu, Mingkun Yang, Zhaohai Li, Jianqiang Wan, Pengfei Wang, Wei Ding, Zheren Fu, Yiheng Xu, Jiabo Ye, Xi Zhang, Tianbao Xie, Zesen Cheng, Hang Zhang, Zhibo Yang, Haiyang Xu, and Junyang Lin.
\newblock Qwen2.5-vl technical report.
\newblock \emph{arXiv preprint arXiv:2502.13923}, 2025.

\bibitem[Chen et~al.(2021)Chen, Tang, Qin, Liang, Liu, Xing, and Lin]{geoqa}
Jiaqi Chen, Jianheng Tang, Jinghui Qin, Xiaodan Liang, Lingbo Liu, Eric~P Xing, and Liang Lin.
\newblock Geoqa: A geometric question answering benchmark towards multimodal numerical reasoning.
\newblock \emph{arXiv preprint arXiv:2105.14517}, 2021.

\bibitem[Chen et~al.(2024)Chen, Liu, Li, An, Deng, Feng, Zhao, and Xie]{reasoning-agent}
Jiaxing Chen, Yuxuan Liu, Dehu Li, Xiang An, Weimo Deng, Ziyong Feng, Yongle Zhao, and Yin Xie.
\newblock Plug-and-play grounding of reasoning in multimodal large language models.
\newblock \emph{arXiv preprint arXiv:2403.19322}, 2024.

\bibitem[Chen et~al.(2023)Chen, Li, Dong, Zhang, He, Wang, Zhao, and Lin]{ShareGPT4V}
Lin Chen, Jisong Li, Xiaoyi Dong, Pan Zhang, Conghui He, Jiaqi Wang, Feng Zhao, and Dahua Lin.
\newblock Sharegpt4v: Improving large multi-modal models with better captions.
\newblock \emph{arXiv preprint arXiv:2311.12793}, 2023.

\bibitem[Chen et~al.(2022)Chen, Ma, Wang, and Cohen]{pot}
Wenhu Chen, Xueguang Ma, Xinyi Wang, and William~W Cohen.
\newblock Program of thoughts prompting: Disentangling computation from reasoning for numerical reasoning tasks.
\newblock \emph{arXiv preprint arXiv:2211.12588}, 2022.

\bibitem[Chia et~al.(2024)Chia, Toh, Ghosal, Bing, and Poria]{puzzle-vqa}
Yew~Ken Chia, Vernon Toh, Deepanway Ghosal, Lidong Bing, and Soujanya Poria.
\newblock Puzzlevqa: Diagnosing multimodal reasoning challenges of language models with abstract visual patterns.
\newblock In \emph{Findings of the Association for Computational Linguistics: ACL 2024}, pages 16259--16273, 2024.

\bibitem[Dai et~al.(2023)Dai, Li, LI, Tiong, Zhao, Wang, Li, Fung, and Hoi]{InstructBLIP}
Wenliang Dai, Junnan Li, DONGXU LI, Anthony Tiong, Junqi Zhao, Weisheng Wang, Boyang Li, Pascale~N Fung, and Steven Hoi.
\newblock Instructblip: Towards general-purpose vision-language models with instruction tuning.
\newblock In \emph{Advances in Neural Information Processing Systems}, pages 49250--49267, 2023.

\bibitem[Hudson and Manning(2019)]{gqa}
Drew~A Hudson and Christopher~D Manning.
\newblock {GQA}: A new dataset for real-world visual reasoning and compositional question answering.
\newblock In \emph{Proceedings of the IEEE/CVF conference on computer vision and pattern recognition}, pages 6700--6709, 2019.

\bibitem[Jiang et~al.(2024)Jiang, Sun, Sourati, Ahrabian, Ma, Ilievski, Pujara, et~al.]{marvel}
Yifan Jiang, Kexuan Sun, Zhivar Sourati, Kian Ahrabian, Kaixin Ma, Filip Ilievski, Jay Pujara, et~al.
\newblock Marvel: Multidimensional abstraction and reasoning through visual evaluation and learning.
\newblock \emph{Advances in Neural Information Processing Systems}, 37:\penalty0 46567--46592, 2024.

\bibitem[Li et~al.(2023)Li, Li, Savarese, and Hoi]{LLM_BLIP2}
Junnan Li, Dongxu Li, Silvio Savarese, and Steven Hoi.
\newblock Blip-2: Bootstrapping language-image pre-training with frozen image encoders and large language models.
\newblock In \emph{International conference on machine learning}, pages 19730--19742. PMLR, 2023.

\bibitem[Lin et~al.(2009)Lin, Wu, Porway, and Xu]{a-sig}
Liang Lin, Tianfu Wu, Jake Porway, and Zijian Xu.
\newblock A stochastic graph grammar for compositional object representation and recognition.
\newblock \emph{Pattern Recognition}, 42\penalty0 (7):\penalty0 1297--1307, 2009.

\bibitem[Liu et~al.(2023{\natexlab{a}})Liu, Li, Li, and Lee]{LLaVa1.5}
Haotian Liu, Chunyuan Li, Yuheng Li, and Yong~Jae Lee.
\newblock Improved baselines with visual instruction tuning.
\newblock \emph{arXiv preprint arXiv:2310.03744}, 2023{\natexlab{a}}.

\bibitem[Liu et~al.(2024)Liu, Li, Li, Li, Zhang, Shen, and Lee]{llava-next}
Haotian Liu, Chunyuan Li, Yuheng Li, Bo Li, Yuanhan Zhang, Sheng Shen, and Yong~Jae Lee.
\newblock Llava-next: Improved reasoning, ocr, and world knowledge, 2024.

\bibitem[Liu et~al.(2025)Liu, Su, and et~al]{kimi}
Jingyuan Liu, Jianlin Su, and Xingcheng~Yao et al.
\newblock Muon is scalable for llm training, 2025.

\bibitem[Liu et~al.(2023{\natexlab{b}})Liu, Duan, Zhang, Li, Zhang, Zhao, Yuan, Wang, He, Liu, et~al.]{MMBench}
Yuan Liu, Haodong Duan, Yuanhan Zhang, Bo Li, Songyang Zhang, Wangbo Zhao, Yike Yuan, Jiaqi Wang, Conghui He, Ziwei Liu, et~al.
\newblock Mmbench: Is your multi-modal model an all-around player?
\newblock \emph{arXiv preprint arXiv:2307.06281}, 2023{\natexlab{b}}.

\bibitem[Lu et~al.(2022)Lu, Mishra, Xia, Qiu, Chang, Zhu, Tafjord, Clark, and Kalyan]{sqa}
Pan Lu, Swaroop Mishra, Tony Xia, Liang Qiu, Kai-Wei Chang, Song-Chun Zhu, Oyvind Tafjord, Peter Clark, and Ashwin Kalyan.
\newblock Learn to explain: Multimodal reasoning via thought chains for science question answering.
\newblock In \emph{The 36th Conference on Neural Information Processing Systems (NeurIPS)}, 2022.

\bibitem[Ma et~al.(2025)Ma, Huang, and Kun~Yan]{step}
Guoqing Ma, Haoyang Huang, and et~al Kun~Yan.
\newblock Step-video-t2v technical report: The practice, challenges, and future of video foundation model, 2025.

\bibitem[Raven(2003)]{rpm}
Jean Raven.
\newblock Raven progressive matrices.
\newblock In \emph{Handbook of nonverbal assessment}, pages 223--237. Springer, 2003.

\bibitem[Sanh et~al.()Sanh, Webson, Raffel, Bach, Sutawika, Alyafeai, Chaffin, Stiegler, Raja, Dey, et~al.]{sft-nlp-multitask}
Victor Sanh, Albert Webson, Colin Raffel, Stephen Bach, Lintang Sutawika, Zaid Alyafeai, Antoine Chaffin, Arnaud Stiegler, Arun Raja, Manan Dey, et~al.
\newblock Multitask prompted training enables zero-shot task generalization.
\newblock In \emph{International Conference on Learning Representations}.

\bibitem[Shao et~al.(2024)Shao, Wang, Zhu, Xu, Song, Bi, Zhang, Zhang, Li, Wu, et~al.]{deepseekmath}
Zhihong Shao, Peiyi Wang, Qihao Zhu, Runxin Xu, Junxiao Song, Xiao Bi, Haowei Zhang, Mingchuan Zhang, YK Li, Y Wu, et~al.
\newblock Deepseekmath: Pushing the limits of mathematical reasoning in open language models.
\newblock \emph{arXiv preprint arXiv:2402.03300}, 2024.

\bibitem[Shi et~al.(2024)Shi, Hu, Bin, Liu, Yang, Ng, Bing, and Lee]{math-llava}
Wenhao Shi, Zhiqiang Hu, Yi Bin, Junhua Liu, Yang Yang, See~Kiong Ng, Lidong Bing, and Roy Lee.
\newblock Math-llava: Bootstrapping mathematical reasoning for multimodal large language models.
\newblock In \emph{Findings of the Association for Computational Linguistics: EMNLP 2024}, pages 4663--4680, 2024.

\bibitem[Singh et~al.(2019)Singh, Natarajan, Shah, Jiang, Chen, Batra, Parikh, and Rohrbach]{TextVQA}
Amanpreet Singh, Vivek Natarajan, Meet Shah, Yu Jiang, Xinlei Chen, Dhruv Batra, Devi Parikh, and Marcus Rohrbach.
\newblock Towards vqa models that can read.
\newblock In \emph{Proceedings of the IEEE/CVF conference on computer vision and pattern recognition}, pages 8317--8326, 2019.

\bibitem[Tan et~al.(2024)Tan, Wei, Gao, Sun, Li, Guo, Yu, and Li]{mc-cot}
Cheng Tan, Jingxuan Wei, Zhangyang Gao, Linzhuang Sun, Siyuan Li, Ruifeng Guo, Bihui Yu, and Stan~Z Li.
\newblock Boosting the power of small multimodal reasoning models to match larger models with self-consistency training.
\newblock In \emph{European Conference on Computer Vision}, pages 305--322. Springer, 2024.

\bibitem[Wang et~al.(2025)Wang, Pan, Shi, Lu, Ren, Zhou, Zhan, and Li]{mathvision}
Ke Wang, Junting Pan, Weikang Shi, Zimu Lu, Houxing Ren, Aojun Zhou, Mingjie Zhan, and Hongsheng Li.
\newblock Measuring multimodal mathematical reasoning with math-vision dataset.
\newblock \emph{Advances in Neural Information Processing Systems}, 37:\penalty0 95095--95169, 2025.

\bibitem[Wang et~al.(2024)Wang, Bai, Tan, Wang, Fan, Bai, Chen, Liu, Wang, Ge, Fan, Dang, Du, Ren, Men, Liu, Zhou, Zhou, and Lin]{Qwen2-VL}
Peng Wang, Shuai Bai, Sinan Tan, Shijie Wang, Zhihao Fan, Jinze Bai, Keqin Chen, Xuejing Liu, Jialin Wang, Wenbin Ge, Yang Fan, Kai Dang, Mengfei Du, Xuancheng Ren, Rui Men, Dayiheng Liu, Chang Zhou, Jingren Zhou, and Junyang Lin.
\newblock Qwen2-vl: Enhancing vision-language model's perception of the world at any resolution.
\newblock \emph{arXiv preprint arXiv:2409.12191}, 2024.

\bibitem[Wang et~al.(2023)Wang, Lv, Yu, Hong, Qi, Wang, Ji, Yang, Zhao, Song, et~al.]{cogvlm}
Weihan Wang, Qingsong Lv, Wenmeng Yu, Wenyi Hong, Ji Qi, Yan Wang, Junhui Ji, Zhuoyi Yang, Lei Zhao, Xixuan Song, et~al.
\newblock Cogvlm: Visual expert for pretrained language models.
\newblock \emph{arXiv preprint arXiv:2311.03079}, 2023.

\bibitem[Wei et~al.(2022)Wei, Wang, Schuurmans, Bosma, Xia, Chi, Le, Zhou, et~al.]{cot}
Jason Wei, Xuezhi Wang, Dale Schuurmans, Maarten Bosma, Fei Xia, Ed Chi, Quoc~V Le, Denny Zhou, et~al.
\newblock Chain-of-thought prompting elicits reasoning in large language models.
\newblock \emph{Advances in neural information processing systems}, 35:\penalty0 24824--24837, 2022.

\bibitem[Wen et~al.(2024)Wen, Liang, Sierra, Luckin, Tong, Liu, Cui, and Tang]{ai4edu}
Qingsong Wen, Jing Liang, Carles Sierra, Rose Luckin, Richard Tong, Zitao Liu, Peng Cui, and Jiliang Tang.
\newblock Ai for education (ai4edu): Advancing personalized education with llm and adaptive learning.
\newblock In \emph{Proceedings of the 30th ACM SIGKDD Conference on Knowledge Discovery and Data Mining}, pages 6743--6744, 2024.

\bibitem[Wu et~al.(2023)Wu, Yin, Qi, Wang, Tang, and Duan]{ChatGPT}
Chenfei Wu, Shengming Yin, Weizhen Qi, Xiaodong Wang, Zecheng Tang, and Nan Duan.
\newblock Visual chatgpt: Talking, drawing and editing with visual foundation models.
\newblock \emph{arXiv preprint arXiv:2303.04671}, 2023.

\bibitem[Xu et~al.()Xu, Fu, Gao, Ye, Liu, Mei, Wang, Yu, and Wu]{dpo-or-ppo}
Shusheng Xu, Wei Fu, Jiaxuan Gao, Wenjie Ye, Weilin Liu, Zhiyu Mei, Guangju Wang, Chao Yu, and Yi Wu.
\newblock Is dpo superior to ppo for llm alignment? a comprehensive study.
\newblock In \emph{Forty-first International Conference on Machine Learning}.

\bibitem[Yu et~al.(2023)Yu, Yang, Li, Wang, Lin, Liu, Wang, and Wang]{MMVet}
Weihao Yu, Zhengyuan Yang, Linjie Li, Jianfeng Wang, Kevin Lin, Zicheng Liu, Xinchao Wang, and Lijuan Wang.
\newblock Mm-vet: Evaluating large multimodal models for integrated capabilities.
\newblock \emph{arXiv preprint arXiv:2308.02490}, 2023.

\bibitem[Zhang et~al.(2019)Zhang, Gao, Jia, Zhu, and Zhu]{raven}
Chi Zhang, Feng Gao, Baoxiong Jia, Yixin Zhu, and Song-Chun Zhu.
\newblock Raven: A dataset for relational and analogical visual reasoning.
\newblock In \emph{Proceedings of the IEEE/CVF conference on computer vision and pattern recognition}, pages 5317--5327, 2019.

\bibitem[Zhang et~al.(2023)Zhang, Zhang, Li, Zhao, Karypis, and Smola]{multimodal-cot}
Zhuosheng Zhang, Aston Zhang, Mu Li, Hai Zhao, George Karypis, and Alex Smola.
\newblock Multimodal chain-of-thought reasoning in language models.
\newblock \emph{arXiv preprint arXiv:2302.00923}, 2023.

\bibitem[Zhou et~al.(2024)Zhou, Zhou, Hu, Lu, Gao, and Zhang]{image-of-thought}
Qiji Zhou, Ruochen Zhou, Zike Hu, Panzhong Lu, Siyang Gao, and Yue Zhang.
\newblock Image-of-thought prompting for visual reasoning refinement in multimodal large language models.
\newblock \emph{arXiv preprint arXiv:2405.13872}, 2024.

\bibitem[Zhu et~al.(2023)Zhu, Chen, Shen, Li, and Elhoseiny]{LLM_MiniGPT4}
Deyao Zhu, Jun Chen, Xiaoqian Shen, Xiang Li, and Mohamed Elhoseiny.
\newblock Minigpt-4: Enhancing vision-language understanding with advanced large language models.
\newblock \emph{arXiv preprint arXiv:2304.10592}, 2023.

\bibitem[Zhu et~al.(2024{\natexlab{a}})Zhu, Wang, Sun, Chen, Liu, Zhang, and Wang]{nsft}
Ke Zhu, Yu Wang, Yanpeng Sun, Qiang Chen, Jiangjiang Liu, Gang Zhang, and Jingdong Wang.
\newblock Continual sft matches multimodal rlhf with negative supervision.
\newblock \emph{arXiv preprint arXiv:2411.14797}, 2024{\natexlab{a}}.

\bibitem[Zhu et~al.(2024{\natexlab{b}})Zhu, Zhao, Ge, and Zhang]{seva}
Ke Zhu, Liang Zhao, Zheng Ge, and Xiangyu Zhang.
\newblock Self-supervised visual preference alignment.
\newblock \emph{arXiv preprint arXiv:2404.10501}, 2024{\natexlab{b}}.

\end{thebibliography}
